\documentclass[a4paper,11pt,numreferences,mathsec,kaplist]{isuimei}
\usepackage{isu}
\usepackage{booktabs}

\begin{document}

\begin{article}
\setcounter{aqwe}{2}

%%%%%%%%%%%%%%%%%%%%%%%%%%%%%%%%%%%%%%%%%%%%%%%%%%%%%%%%%%%%%%%%%%%%%%%%%%%%%%%
\begin{opening}
%пїЅпїЅпїЅ
\udk{518.517}
% пїЅпїЅпїЅпїЅпїЅпїЅпїЅпїЅ пїЅпїЅпїЅпїЅпїЅпїЅпїЅ пїЅ пїЅпїЅпїЅпїЅпїЅпїЅпїЅпїЅпїЅпїЅпїЅпїЅпїЅ
\title{Power System Parameters Forecasting Using \\ Hilbert-Huang Transform and Machine Learning\thanks{This work was supported by the grant of Leading Scientific Schools of Russian Federation No.1507NSH-1507.20128. D. N. Sidorov and V. A. Spiryaev are supported by  grant of RFBR No. 12-01-00722. N. V. Tomin is supported by the Humboldt Research Fellowship programme of Alexander von Humboldt Foundation (Germany).  P. Leahy is supported by the Stokes Lectureship programme of Science Foundation Ireland. D. N. Sidorov is supported by Russian Federal framework programme, state contract No.14.B37.21.0365 (Russia).}}
%пїЅпїЅпїЅпїЅпїЅпїЅ
\author{V.G.~\surname{Kurbatsky}, V.A.~\surname{Spiryaev}, N.V.~\surname{Tomin}}
%пїЅпїЅпїЅпїЅпїЅ пїЅпїЅпїЅпїЅпїЅпїЅ
\institute{Energy Systems Institute, Siberian Branch of Russian Academy of Sciences}
\author{P.~\surname{Leahy}}
\institute{University College Cork}
\author{D.N.~\surname{Sidorov}, A.V.~\surname{Zhukov}}
%пїЅпїЅпїЅпїЅпїЅ пїЅпїЅпїЅпїЅпїЅпїЅ
\institute{Energy Systems Institute, Siberian Branch of Russian Academy of Sciences\\ Institute of Mathematics, Economics and Informatics, Irkutsk State University}

% пїЅпїЅпїЅпїЅпїЅпїЅпїЅпїЅпїЅпїЅ пїЅпїЅпїЅ пїЅпїЅпїЅпїЅпїЅпїЅпїЅпїЅпїЅпїЅпїЅ
\runningtitle{Power System Parameters Forecasting}
\runningauthor{V.G.~Kurbatsky et al}

% пїЅпїЅпїЅпїЅпїЅ пїЅпїЅпїЅпїЅпїЅпїЅпїЅпїЅпїЅ
\begin{abstract}

 A novel hybrid data-driven approach is developed for forecasting power system para\-meters with the goal of increasing the efficiency of short-term forecasting studies for non-stationary time-series. The proposed approach is based on mode decomposition and a feature analysis of initial retrospective data using the Hilbert-Huang transform and machine learning algorithms. 
The random forests and gradient boosting trees learning techniques were examined.  The decision tree techniques were used to rank the importance of variables employed in the forecasting models.  The Mean Decrease Gini index is employed as an impurity function. 
The resulting hybrid forecasting models employ the radial basis function neural network and support vector regression.

Apart from introduction and references the paper is organized as follows. The second section  presents the background and the review of several approaches for short-term forecasting of power system parameters. In the third section a hybrid machine learning-based algorithm using Hilbert-Huang transform is developed for short-term forecasting of power system parameters. Fourth section 
describes the decision tree learning algorithms used for the issue of variables importance.
Finally in section six the experimental results in the following electric power problems are presented: active power flow forecasting, electricity price forecasting and for the wind speed and direction forecasting.

\end{abstract}

%\keywords{пїЅпїЅпїЅпїЅпїЅпїЅпїЅпїЅпїЅ пїЅпїЅпїЅпїЅпїЅ 5 пїЅпїЅпїЅпїЅпїЅпїЅпїЅпїЅ пїЅпїЅпїЅпїЅ.}

\end{opening}
%%%%%%%%%%%%%%%%%%%%%%%%%%%%%%%%%%%%%%%%%%%%%%%%%%%%%%%%%%%%%%%%%%%%%%%%%%%%%%%%%%%%%%%%%%%

%пїЅпїЅпїЅпїЅпїЅпїЅпїЅпїЅпїЅпїЅ пїЅпїЅпїЅ пїЅпїЅпїЅпїЅпїЅпїЅпїЅпїЅпїЅпїЅ
\avtogl{V.G. Kurbatsky, N.V.Tomin, V.A.Spiryaev, P.Leahy, D.N.Sidorov}{Power System Parameters Forecasting Using  Hilbert-Huang Transform and Machine Learning}

% пїЅпїЅпїЅпїЅпїЅ пїЅпїЅпїЅпїЅ пїЅпїЅпїЅпїЅпїЅ пїЅпїЅпїЅпїЅпїЅпїЅ

\section{ Introduction }

In recent years power industries worldwide have experienced two major changes: liberalization of the electricity market and the expansion of renew\-able energy. Liberalization has led to a tremendous increase in the inter-regional wholesale electricity trade between neighboring utilities or regions. The increasing penetration of renewable energy has led to increasing depen\-dency on the volatile nature of renewable energy sources. Because of this, changes in a power system's parameters often feature a sharply variable non-stationary behavior, which limits the efficiency of modern technologies for forecasting time series \cite{lit1, lit2}.

In this situation, effective short-term forecasting of state variables is highly important for market participants in wholesale power markets and for the power system as a whole. Over the past decade, many power engineers have focused their efforts on novel prediction problems, especially short-term forecasting of power flow, wind power forecasting and electricity price forecasting. In addition to difficulties with the sharply variable non-stationary behavior of power system parameters, researchers deal with a data requireпїЅ-ments problem in actual forecasting of such parameters. The required histori\-cal data is either not publicly available on power companies' websites or only made available with some delay. Consequently, the correct pattern of possible future variability of power system parameters is difficult to effectively identify, which in turn prevents accurate forecasting of system variables.

In this paper a novel hybrid approach for power system parameters forecasting is suggested, which combines the effective apparatus of analysis of non-stationary time series -- the Hilbert-Huang integral transform (HHT) and machine learning technologies. 

The paper is organized as follows. Section 2 presents the background and the review of several approaches for short-term forecasting of power system parameters. In section 3 a hybrid machine learning-based algorithm using HHT is developed for short-term forecasting of power system parameters. Section 4 presents the experimental results and discussion.

\section{ Background}

In the electricity grid at any moment balance must be maintained between electricity consumption and generation - otherwise disturbances in power quality or supply may occur. Forecasting power systems parameters may be considered at different time scales, depending on the intended application. From milliseconds up to a few minutes, forecasts can be used, for instance, for turbine active control. This type of forecast is usually referred to as very short-term forecasts. For the following 48--72 hours, forecasts are needed for power system management or energy trading.

The short-term forecasting of power system parameters can be carried out both with the aid of classical approaches of dynamic estimation, statisti\-cal methods of analysis of time series and regressive models, and with the aid of artificial intelligence. Many techniques have been employed for such purposes, including machine learning techniques -- artificial neural networks (ANNs) \cite{lit3, lit4}, support vector machines (SVMs) \cite{lit5}, random forest models 
\cite{lit5, lit6} and etc. Moreover, time series models, like ARIMA, GARCH models \cite{lit7, lit8}, Kalman filter-based algorithm \cite{lit9, lit10} have also been proven to be effective in the power system parameters forecasting. 

It is worth noting that power systems parameters often feature sharply variable non-stationary behaviour, which limits the efficiency of the stated technologies for forecasting of time series. Studies have shown that hybrid approaches, that is a combination of different intelligent techniques, have great potential and are worth pursuing \cite{lit2}. Previously, several hybrid fore\-casting approaches have been proposed, including ARIMA-ANN models \cite{lit11}, Fuzzy-ANN-based models \cite{lit12}, Wavelet-ANN-based models \cite{lit13}, Fuzzy-expert sy\-stem-based models \cite{lit14} among others \cite{lit15}. For review and biblio\-graphy of the state-ofthe-art medods for dynamical systems identification and fore\-casting readers may refer to the monograph  \cite{lit19_1}.

It is well known that for effective forecasting of power system parameters a number of additional, correlated characteristics must be taken into account, which can be influenced by a predictable parameter. For instance, everything from salmon migration to forest fires can affect current and future electricity prices. But quite often, information about these characteristics is either unavailable or limited and contradictory. As a result, scientific researchers and power engineers have to work with only one retrospective predictable parameter time series without additional characteristics, which imposes limitations on forecasting performance. We consider that, in such a case,  forecasting performance can be improved by using a transition from a time series task to a regression task, when an initial time series is decomposed into some components.  An investigation of properties of obtained components can help to improve the performance of a power system parameter forecast.

\section{ Proposed Approach}

We propose a hybrid machine learning-based algorithm using HHT as a data preprocessing technology \cite{lit16,lit17}. Figure 1 the shows general forecasting scheme. There are five main blocks.

Data collection is performed by the SCADA system, which merely identi\-fies a set of power system parameters. The second block preprocesses input data with HHT. HHT is a two-step algorithm; combining empirical mode decomposition (EMD) and the Hilbert transform (HT). After the HHT was employed we obtain, through EMD and HT, the sets of intrinsic mode functions (IMFs), instantaneous amplitudes and frequencies. These sets are used as input values to the next block for feature selection and dimensionality reduction. In this block, learning methods are used -- Random Forest (RF) and Boosting Trees (BT) to rank a variable's importance \cite{lit18,lit19}. SVM and ANN forecasting models with non-linear optimization algorithms are employed in the next block for optimal selection of parameters. The last block is used for testing and a comparison of obtained forecasting models. As a result, an HHT-ANN-SVM-based forecasting system will be able to predict power system parameters.

\begin{figure}[htbp]
	\centering
		\includegraphics[scale=0.5]{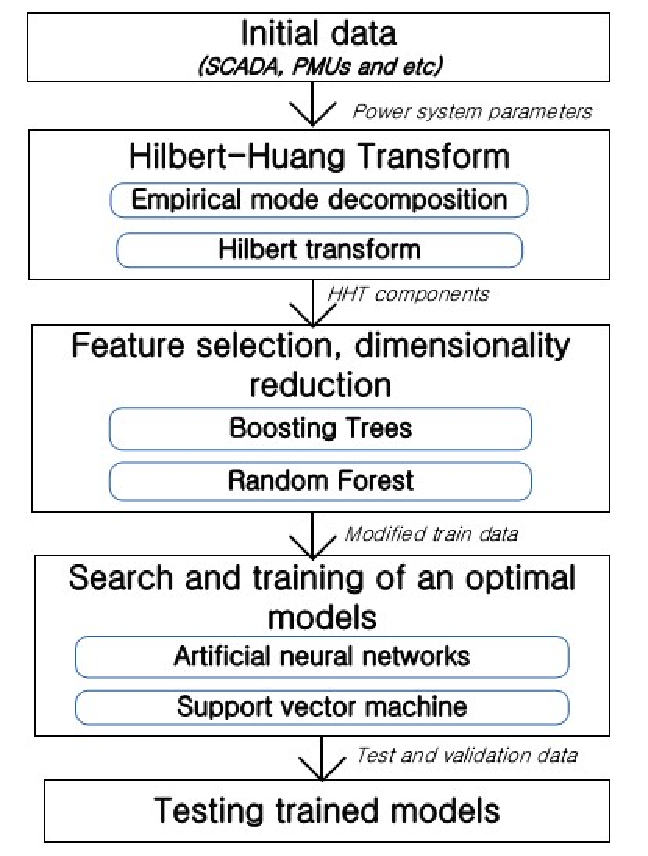}
		\caption{Block diagram of the proposed hybrid approach scheme.}
		%Fig. 1. Block diagram of the proposed hybrid approach scheme.
	\label{fig:image1}
\end{figure}

The main goal of the hybrid forecasting approach development was to increase the efficiency of forecasting studies, when we deal with limited retrospective data and sharply variable, non-stationary time series of power system parameters.     

%\section{Machine learning techniques to power systems parameters forecasting}

\section{Decision tree learning algorithms for the issue of variables importance}

Forecasting of wind power, power flows, energy price etc. have become a major issue in power systems. Following the needs of today's competitive environment, various techniques are used to forecast wind power, energy price and power demand. Over the past decade, different machine learning algorithms have been investigated for analyzing power systems parameters' patterns in order to obtain forecasts in power systems with high accuracy.  

%\subsection{Decision tree learning algorithms for variables importance issue}
The importance of a variable is the contribution it makes to the success of the model. For a predictive model, success means good forecasting. Often the forecasting relies mainly on a few variables. A good measure of importance reveals those variables. The better the forecasting, the closer the model represents reality, and the more plausible it is that the important variables represent the true cause of forecasting.

We used the decision tree techniques to rank the importance of variables.
These algorithms can output a list of predictor variables that they believe to be important in predicting the outcome. If nothing else, researchers can identify a subset of the data to only include the most ``important'' variables, and use that with another model. 

As it is explained in \cite{aleksei1}, every binary decision tree is separately represented by a tree structure T, from an input vector $(X_{1},...,X_{p})$ taking its values in $(\chi_{1}*...*\chi_{p}=\chi)$ to a output variable $y \in Y$. Every certain node t represents a subset of the space $\chi$, with the root node being $\chi$ itself. Construction of decision trees usually works top-down, by choosing a variable at each step that splits the set of items by the binary test $s_{t}=(X_{m}<c)$. The internal node t divides its subset into two subsets\footnote{More generally, splits are defined by a (not necessarily binary) partition of the range $\chi_{m}$ of possible values of a single variable $\chi_{m}$} corresponding to two children nodes $t_{L}$ and $t_{R}$. For a new instance the predicted value $\hat{Y}$ is the label of the leaf reached by the instance when it is propagated through the tree. Algorithms for constructing decision trees identify at each node t the split $s_{t} = s^{*}$ for which the partition of the $N_{t}$ node samples into $t_{L}$ and $t_{R}$ maximizes the decrease
$$
\Delta i(s,t)=i(t) - p_{L}i(t_{L}) - p_{R}i(t_{R})
$$
With regard to variable importance in Random forests, in \cite{lit19} it was proposed to add up the weighted impurity (loss) decreases $p(t)i(s_{t},t)$ for all nodes t where $X_{m}$ is used, averaged over all $N_{T}$ trees in the forest:
$$
Imp(X_{m}) = \frac{1} N_{t} \sum_{T}\sum_{t\epsilon T:\nu(s_{t}) = X_{m}}p(t)i(s_{t},t)
$$
and where $p(t$) is the proportion $\frac{N_{t}}N$ of samples reaching t and $\nu(s_{t})$ is a variable which is used in split $s_{t}$.

In our paper we have examined two decision tree learning techniques: random forests and gradient boosting trees. We used the decision tree techniques to rank the importance of variables. Also we use Gini index, which is also called Mean Decrease Gini, as an impurity function. It was introduced for decision trees in \cite{nikita2} and later modified for gradient boosting machines \cite{nikita3}.

\section{Case Study}

The novel hybrid approach is realized in STATISTICA 6.0. and Matlab. The efficiency of the developed approach is demonstrated with real time series in the following electric power problems: active power flow forecasting for the Western Baikal-Amur Mainline (Russia) 
\cite{lit16,lit20}, electricity price forecasting for the Australian and Nord Pool Spot (EU) power markets
\cite{lit17}, wind speed fore\-casting for the Valentia region (Ireland) \cite{lit10}.

\subsection{Valentia wind speed forecasting (Ireland)}

\paragraph{HHT-ANN-SVM model}

The initial training data is represented by two per-hour time-series: wind speed and wind direction values for 1 year for the Valentia region. Both time series were decomposed into IMFs by the Huang method, and the Hilbert transform was employed to obtain the amplitudes, $A$ and frequency, $F$. The latter along with these HHT components were used as input values of the selected HHT-ANN and HHT-SVM models. The final 48 hours of the data were reserved for calculating performance statistics. 

Fig. 2 illustrates the feature selection analysis of the modified Valentia wind speed data using regression machine learning algorithms -- Boosting Trees and Random Forest. As seen, high-frequency components of the initial wind speed data (IMF1, IMF2, $A_1$, $A_2$ and $F_1$, $F_2$) have less predictive importance (less than 0.3) by comparison with other components, and therefore they can be excluded from the trained data.

\begin{figure}[htbp]
	\centering
		\includegraphics[scale=1]{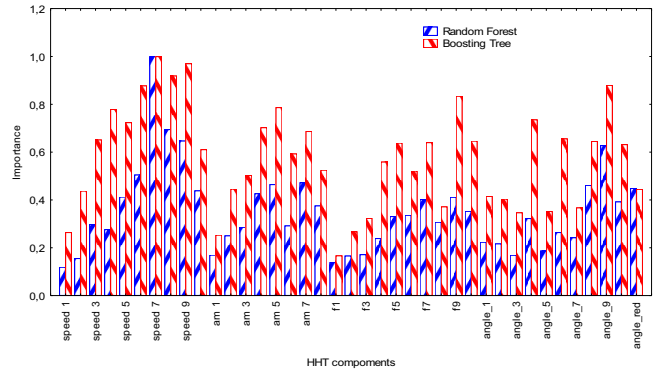}
		\caption{Predictor importance diagram based on Random Forest and Boosting Trees learning algorithms for the Valentia wind speed data.}
		%Fig. 2. Predictor importance diagram based on Random Forest and Boosting Trees learning algorithms for the Valentia wind speed data.
	\label{fig:image2}
\end{figure}

As a result, the following hybrid models are constructed, trained and tested: a RBF neural network (HHT-ANN model) and epsilon-SVM regression model (HHT-SVM model). The RBF network is trained by a 2-Phase-RBF scheme (two-step algorithm). After a training-testing process we obtained the following errors: train error -- 0.251\%, validation error -- 0.263\% and test error -- 0.257\%. The obtained RBF network had the following structure: input neurons 25, hidden neurons 162 and output neuron 1. 

The polynomial kernel function is used in the  epsilon-SVM regression model. Let us briefly outline the SVM regression model.
Let $(x_i, y_i),\, i=\overline{1,N},$ be the input data, namely $x_i \in \widetilde{X} \subset X, y_i \in \widetilde{Y} \subset Y,$ $X$ is the set of input data, $Y$ is the set of values of the desired objective function. SVM resolves the regression problem by estimation of the unknown
function 
$$
y_i = f(x_i)+\varepsilon_i, 
\eqno(1)
$$
 where $\varepsilon_i$ is error of  $i$-th observation, e.g. in linear case
$
f({\bf x}) = \left<{\bf w},{\bf x} \right>+b,
$
 where ${\bf w}\in\mathbb{R}^n$ are weights and     $b\in\mathbb{R}$ is the offset parameter.
In nonlinear case SVM employs $\phi: \mathbb{R}^n \rightarrow H$  and instead of 
 (1) the following function $f$ is under the estimation
$f({\bf x}) = \left<{\bf w},\phi({\bf x})\right>+b.$
 Determination of parameters is reduced to a convex optimization problem, i.e. the the following regularized risk functional minimization
%\begin{equation}
$$
\min\limits_{{\bf w},b, \xi,\xi^*} \left[ \frac{1}{2}||{\bf w}||^2 +C\sum\limits_{i=1}^N(\xi_i +\xi_i^*)  \right],
\eqno(2)
$$
%\label{svm1}
%\end{equation}
% \begin{equation}
$$
 \left\{ \begin{array}{ll}
         \mbox{$y_i - \left< {\bf w, \phi(x_i)} \right>- b\leq \varepsilon +\xi_i $}, \\
         \mbox{$\left< {\bf w, \phi(x_i)} \right>+ b - y_i\leq \varepsilon +\xi_i^* $}, \\
         \mbox{$\xi_i, \xi_i^* \geq 0,\, i=\overline{1,N}.$} 
        \end{array} \right. 
\eqno(3)
$$
%\label{svm2} 
  %       \end{equation}
Here  $C>0$ is the capacity constant to trade-off between the flatness
of $f$ and the amount up to which deviations larger than
$\varepsilon$ are tolerated, $\xi_i$ are  inseparable observations parameters, $\varepsilon$ is losses parameter.   The second term in the functional (2) penalizes any deviation $f(x_i)$ from $y_i$ based on the constraints (3). This problem is resolved in terms of 
the dual problem to the problem (2) -- (3) as follows
$
f(x) = \sum\limits_{i=1}^N (\alpha_i - \alpha_i^*)K(x_i,x)+b,
$
where
 $0\leq \alpha_i, \alpha_i^* \leq C, $ and kernel $K(x_i,x)$
satisfies the conditions of Mercer-Hilbert-Schmidlt theorem (symmetry and non-negativity \cite{merc_lit})
and can be represented as follows $K(x_i,x) = \left< \phi(x_i), \phi(x) \right >,$ which is the RBF
is our case.
 Using a 10-fold cross-validation method the following optimal training constants and RBF kernel parameters were determined: the capacity constant $C=1$; $\varepsilon=1.3$; $d=3$;$\gamma=0.03$.

\paragraph{ FF-ANN-SA model}

In this case example a feed forward ANN is tested for performance comparison with the hybrid HHT approach. The candidate input vectors for the FF-ANN-SA model were: the current value of wind speed at time $t$; previous values of wind speeds at times $(t - 1)$,  $(t - 3)$, $(t - 6)$; wind direction at times $t$ and $(t - 1)$; and time of day. All input vectors and the target vector were normalized to the range $[0,1]$, and the data were divided in the ratio 4:1 between training and validation sets, with the final 48 hours reserved for testing as before. 

A hybrid simulated annealing/tabu search algorithm for simultaneous optimization of the network structure and weights was applied to the problem of 24h ahead prediction of wind speeds [21, 22]. The structural optimization element of the algorithm uses a network pruning approach, which involves defining an initial, maximal network and then randomly removing intra-neuron connections until a more parsimonious network is arrived at (Fig. 3). The inter-neuron connectivity and weights were optimized using the hybrid algorithm prior to subsequent optimization of weights and biases using Levenberg-Marquadt backpropagation.

\begin{figure}[htbp]
	\centering
		\includegraphics[scale=1.38]{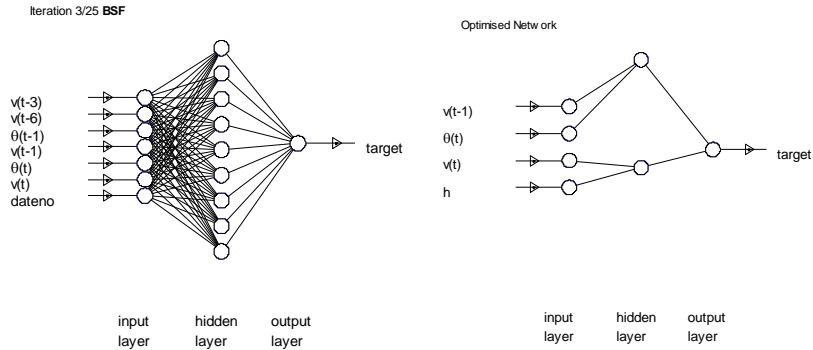}
		\caption{(a) maximal network; (b) FF-ANN-SA network with optimised structure.}
		%Fig. 3. (a) maximal network; (b) FF-ANN-SA network with optimised structure.
	\label{fig:image3}
\end{figure}

The maximal network was defined as having all seven candidate inputs, and a single hidden layer with nine neurons. Therefore, the maximum number of inter-neuron connections was $N_1 N_2 + N_2 N_3$ where $N_1$ is the number of inputs, $N_2$ the number of hidden layer neurons, and $N_3$ the number of outputs. The cost function specified for the optimization is based on predictive performance and number of inter-neuron connections in the network, with a 3:1 weighting ratio between network complexity and per\-for\-mance. Tan-sigmoidal and linear activation functions were used for the hidden and output layer neurons, respectively. The linear activation function of the output layer was constrained to produce only positive outputs, in order to represent the physical reality of wind speeds which are zero-limited. The algorithm was allowed to run for a maximum of 25 iterations, or until the GL5 early stopping criterion was satisfied \cite{lit23}.
\begin{figure}[htbp]
	\centering
		\includegraphics[scale=1]{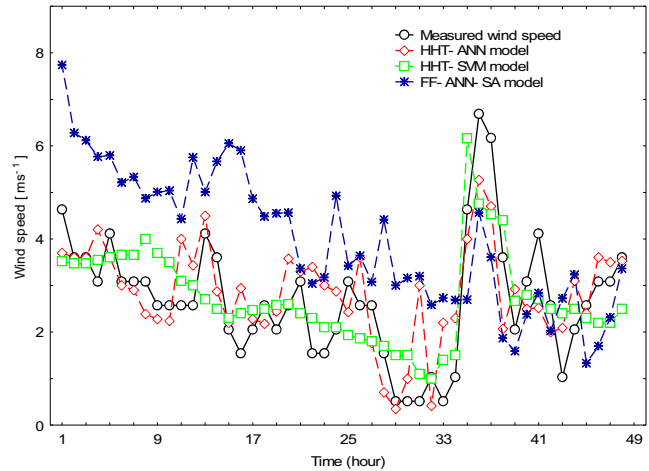}
		\caption{``24-hour ahead'' wind speed forecast based on hybrid HHT-ANN and HHT-SVM models.}
		%Fig. 4. ``24-hour ahead'' wind speed forecast based on hybrid HHT-ANN and HHT-SVM models.
	\label{fig:image4}
\end{figure}
In Table I and Figure 4 the testing results for a ``24 hour ahead'' wind speed forecasting are presented based on the HHT --ANN, HHT--SVM and FF-ANN-SA models. As shown in Table I, the most accurate forecast was given by the hybrid HHT-SVM and HHT-ANN models (MAE for 0-24h period -- 0.58 ms and 0.55 ms respectively), compared with the conven\-tional approaches: the FF-ANN-SA model -- 2.70 ms.
\begin{center}
\begin{table}
\centering
\begin{center}
\hspace*{-2.3cm}
\begin{tabular}{p{0.1in}p{0.5in}p{0.61in}p{0.5in}p{0.6in}p{0.7in}} \toprule
\textbf{пїЅ} & \textbf{Test data\newline interval} & \textbf{Error\newline ms} & \multicolumn{3}{p{1.4in}}{\textbf{Models}} \\ %\hline 
\multicolumn{3}{p{1.4in}}{\textbf{}}& \textbf{\textit{HHT-ANN}} & \textbf{\textit{HHT-SVM}} & \textbf{\textit{FF-ANN-SA}} \\  \midrule
1 & 0-24h & MAE & 0.55 & 0.58 & 2.70 \\ %\hline 
 &  & RMSE & 0.75 & 0.88 & 3.30 \\ %\hline 
2 & 24-48h & MAE & 0.73 & 0.93 & 1.54 \\ %\hline 
 &  & RMSE & 1.05 & 1.17 & 1.69\\ \bottomrule
\end{tabular}
\end{center}
\caption{Comparison of ``24hour ahead'' Wind Speed Forecasts on the Basis of Hybrid and Conventional Neural Approaches for two subsets of the test data.}
\end{table}
\end{center}

The FF-ANN-SA algorithm did not perform as well as the HHT methods but identified the inputs with the most predictive power as: the current wind speed, the wind speed of the previous hour, the wind direction and the time of day.

\subsection{Active power flow forecasting for the Western Baikal-Amur Mainline (Russia)}

The BAM external power supply is carried on the 110 kV power network (from ``Taishet'' to ``Lena'' substations) and the 220 kV power network (from `` Lena'' to ``Taksimo'' substations). Continuous and random changes in time traction load and the presence of a two-way feed greatly complicates  the forecasting of state variables.

\begin{figure}[htbp]
	\centering
		\includegraphics[scale=1]{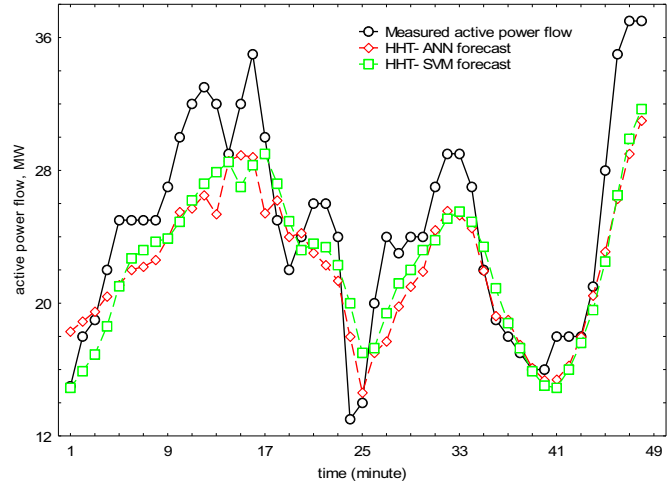}
    \caption{Result of ``1 minute ahead'' active power flow forecasting based on hybrid genetic models.}
	\label{fig:image5}
\end{figure}

The initial training data is represented by the telemetric data of the per-minute active power flow for 3 days (4320 values). The initial time series were converted using the HHT algorithm. The obtained HHT components are used as input values of the trained machine learning-based forecasting models. The MLP neural network model is trained with the Broyden--Fletcher--Goldfarb--Shanno algorithm \cite{lit24}. The radial basis function kernel function is used in the epsilon-SVM regression model. Using a 10-fold cross-validation method the following optimal training constants and kernel para\-meters were determined: the capacity constant, $C=10$; epsilon, $\varepsilon=0.4$; degree, $\gamma=0.001$. To compare the performance of forecasting approaches, the authors also calculated similar forecasts based on conventional ap\-proaches: Auto\-regressive integrated moving average (ARIMA), single ANN and Multi\-plicative Exponen\-tial Smoothing.  

Table II and Fig. 5 summaries the numerical results for a ``1 minute ahead'' active power flow forecasting on based on the HHT-ANN and HHT-SVM models.  As Table II shows, the most accurate forecast was given by the hybrid HHT- SVM and HHT-ANN models (MAPE - 11.6\% and 12.5\% respectively), compared with the conventional approaches: the ANN model -- 15,8\%,  the ARIMA model -- 22.4 \% and Exponential Smoothing -- 32.2\%. 
\begin{table}
%\noindent TABLE II. Comparison of ``1 minute ahead''  Active Power Flow Forecasts on the Basis of Hybrid and Conventional Approaches.
\begin{center}
\hspace*{-4cm}\begin{tabular}{p{0.2in}p{0.8in}p{0.5in}p{0.4in}p{0.5in}} \toprule
\textbf{пїЅ} & \textbf{Forecasting models} & \multicolumn{3}{p{1.4in}}{\textbf{Error}} \\ %\hline 
 &  & \textit{MAPE, \%} & \textit{MAE, MW} & \textit{RMSE}, $MW^2$ \\ \midrule
1 & HHT-ANN & 12,5 & 3.2 & 3,9 \\ %\hline 
2 & HHT-SVM & 11,6 & 2.9 & 3,5 \\ %\hline 
3 & ANN (MLP) & 15,8 & 6,9 & 33,6 \\ %\hline 
4 & ARIMA & 22.4 & 7.1 & 7.4 \\ %\hline 
5 & Exponential Smoothing & 34.2 & 10.8 & 10.9 \\ \bottomrule
\end{tabular}
\end{center}
\caption{Comparison of a ``1 minute ahead''  Acive Power Flow Forecasts on the Basis a Hybrid and Conventional Approaches.}
\end{table}
\subsection{ Electricity price forecasting for the Australian market (Australia)}
The Australian NEM is a duopoly with a dominant player, resulting in electricity price changes associated with the behavior of dominant players, and is very difficult to predict in the short term. To assess the forecasting performance of different models, each dataset is divided into two samples for training and testing. The training dataset is used exclusively for model development, and then the test sample is used to evaluate the established model. To build the forecasting model for each of the considered weeks, the information available includes hourly price historical data of the four weeks prior to the first day of the week whose prices are to be predicted.

Here we present the comparison of forecast performance with the hybrid ARIMA-ANN model proposed by Phatchakorn Areekul et al. in \cite{lit11} and our HHT-ANN hybrid using the data of Australian national electricity market, New South Wales. By analogy with \cite{lit11} for the sake of a fair comparison, the fourth week in one season is selected.

The RF-BT-based importance analysis is shown that in contrast to pre\-vious results, in this case study, two first high-frequency components of electricity price data IMF1 (BT -- 1.0, RF -- 0,79) and IMF2 (BT -- 0.95, RF -- 0,70) have high prediction importance. Components IMF5 and F2 have less prediction importance (less than 0.20) and therefore they can be excluded from the trained data.   

Table III and Figure 6 present the values of RMSE, MAE, and MAPE from the price forecasting performance of the proposed hybrid HHT-ANN-SVM model and hybrid ARIMA-ANN model. The results demonstrate that the proposed HHT-ANN (MAPE -- 12.24\%) hybrid model could provide a considerable improvement of the electricity price forecasting accuracy comparing to the ARIMA-ANN -- 13.03\% and the HHT-SVM -- 17.16\% hybrid models. It is fair to note that the HHT-SVM displays smaller errors indicated by MAE and RMSE (respectively 7.01\% and 23.74\%) compared to the ARIMA-ANN model (7.12\% and 28.02\%).

\begin{figure}[htbp]
	\centering
		\includegraphics[scale=1]{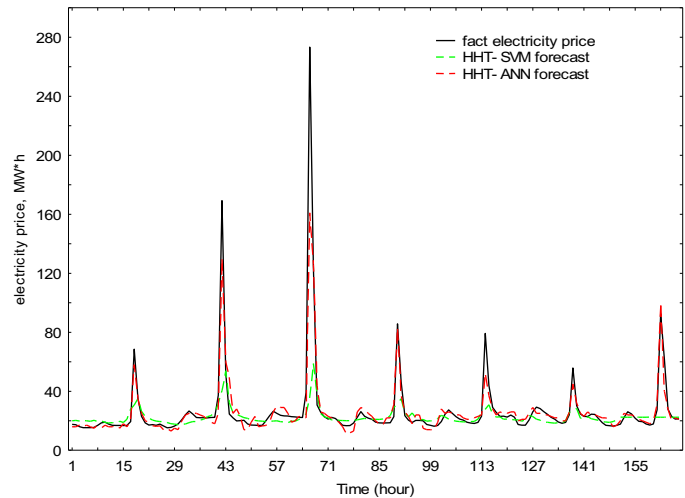}
    \caption{Forecasting results (in AUD/MWh) of May 21-27, 2006 based on hybrid HHT-ANN-SVM model.}
	\label{fig:image6}
\end{figure}

\begin{table}
% \noindent TABLE III.  Comparison of a ``1 hour ahead''  Electricity Price Forecasting on the Basis of Hybrid Approaches.

\begin{center}
\hspace*{-4cm}\begin{tabular}{p{0.2in}p{0.8in}p{0.5in}p{0.4in}p{0.5in}} \toprule
\textbf{пїЅ} & \textbf{Forecasting models} & \multicolumn{3}{p{1.4in}}{\textbf{Error}} \\ %\hline 
 &  & \textit{MAPE, \%} & \textit{MAE, MW} & \textit{RMSE}, $MW^2$ \\ \midrule 
1 & HHT-ANN & 12,24 & 3.86 & 20,12 \\ %\hline 
2 & HHT-SVM & 17,16 & 7.16 & 23,74 \\ %\hline 
3 & ARIMA-ANN & 13,03 & 7,12 & 28,02 \\ \bottomrule
\end{tabular}
\end{center}
\caption{Comparison of ``1 hour ahead''  Electricity Price Forecasting on the Basis of Hybrid Approaches.}
\end{table}

\section{ Conclusions}

In this paper we studied the problem of short-term forecasting of power system parameters. The main contribution is the hybrid machine learning-based algorithm using HHT as a data preprocessing technology. The proposed approach is based on mode decomposition and a feature analysis of initial retrospective data using HHT and BT-RF-based machine learning algorithms for a prediction importance evaluation. We employed machine learning tech\-nologies such as ANNs and SVM as forecasting models.  

Forecast performance comparison studies with other conventional, machine learning and hybrid approaches were also presented, especially with the FF-ANN-SA approach. The hybrid ANN-SA optimization method does not approach the performance of the HHT methods, as indicated by higher RMSE and MAE. Performance of the hybrid optimization ANN method can probably be improved with some further adjustments, but it is not expected that it will approach the performance of the superior HHT methods due to the non-stationarity of the wind speed time series. Also one more circumstance should be noted. As for today there are a lot of results in the HHT time series processing field which are associated with various modifications of the HHT. In \cite{emdadd1}, a modification of the standard approach to the decomposition of IMF was suggested, and in \cite{emdadd2} there is the variation of this proposed modification.
Both of these approaches are based on the idea of adding a white noise to the original signal and the construction of an ensemble of different sets of components.
Averaging over the components of an ensemble provides a set of IMF, which contains the components reflecting the real components of the investigated process more accurately than the original EMD. These modifications of the proposed hybrid machine learning-based algorithm
will be employed in our future works.

\bigskip

\end{article}

\end{document}